\documentclass{article}
\usepackage{ijcai24}
\usepackage{arydshln}
\usepackage{amssymb}  
\usepackage{amsfonts}
\usepackage{graphicx}  
\usepackage[table,xcdraw]{xcolor} 

\definecolor{lightblue}{rgb}{0.88, 0.92, 0.97}

\usepackage{times}
\usepackage{soul}
\usepackage{url}
\usepackage[hidelinks]{hyperref}
\usepackage[utf8]{inputenc}
\usepackage[small]{caption}
\usepackage{graphicx}
\usepackage{amsmath}
\usepackage{amsthm}
\usepackage{booktabs}
\usepackage{algorithm}
\usepackage{algorithmic}
\usepackage[switch]{lineno}

\title{DOEI: Dual Optimization of Embedding Information for Attention-Enhanced Class Activation Maps}

\author{
Hongjie Zhu$^1$,~
Zeyu Zhang$^{2\dag}$,~ %
Guansong Pang$^3$,~
Xu Wang$^4$,~
Shimin Wen$^1$,~\\
Yu Bai$^1$,~
Daji Ergu$^1$,~ %
Ying Cai$^{1}$\thanks{Corresponding author (caiying@swun.edu.cn). $^\dag$Project lead.},~
Yang Zhao$^5$\\ %
\affiliations
$^1$Southwest Minzu University~
$^2$The Australian National University\\
$^3$Singapore Management University~
$^4$Sichuan University~
$^5$La Trobe University
}

\begin{document}

\maketitle

\begin{abstract}

    Weakly supervised semantic segmentation (WSSS) typically utilizes limited semantic annotations to obtain initial Class Activation Maps (CAMs). However, due to the inadequate coupling between class activation responses and semantic information in high-dimensional space, the CAM is prone to object co-occurrence or under-activation, resulting in inferior recognition accuracy. To tackle this issue, we propose \textbf{DOEI}, Dual Optimization of Embedding Information, a novel approach that reconstructs embedding representations through semantic-aware attention weight matrices to optimize the expression capability of embedding information. 
    Specifically, DOEI amplifies tokens with high confidence and suppresses those with low confidence during the class-to-patch interaction. This alignment of activation responses with semantic information strengthens the propagation and decoupling of target features, enabling the generated embeddings to more accurately represent target features in high-level semantic space. 
    In addition, we propose a hybrid-feature alignment module in DOEI that combines RGB values, embedding-guided features, and self-attention weights to increase the reliability of candidate tokens. 
    Comprehensive experiments show that DOEI is an effective plug-and-play module that empowers state-of-the-art visual transformer-based WSSS models to
    significantly improve the quality of CAMs and segmentation performance on popular benchmarks, including {\it PASCAl VOC} (+3.6\%, +1.5\%, +1.2\% mIoU) and {\it MS COCO} (+1.2\%, +1.6\% mIoU). 
    Code will be available at \url{https://github.com/AIGeeksGroup/DOEI}.
    
\end{abstract}

\section{Introduction}

    \begin{figure}[t]
    \centering
    \includegraphics[width=\columnwidth]{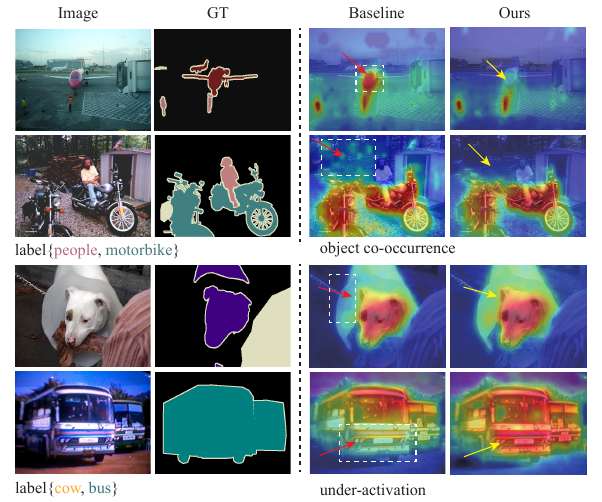}  
    \caption{Visualization comparison between the baseline (MCTformer~\protect\cite{xu2022multi}) and our method. The baseline model results exhibit background noise (object co-occurrence) and less precise localization (under-activation), as highlighted by red arrows and white bounding boxes. In contrast, our proposed method effectively reduces background activation and enhances focus on target regions, as highlighted by yellow arrows.}
    \label{Figure 1}
    \end{figure}

    Semantic segmentation \cite{ge2024esa,zhang2025gamed,wu2023bhsd,zhang2023thinthick} aims to assign each pixel in an image a specific category label~\cite{li2022class,zhang2024segreg,tan2024segstitch,tan2024segkan}. Traditional methods often rely on extensive and precise pixel-level annotations to promote network performance. However, obtaining such annotations is notoriously time-consuming and resource-intensive~\cite{cheng2023out}. Consequently, researchers have increasingly adopted alternative forms of weak supervision, such as scribbles~\cite{vernaza2017learning}, bounding boxes~\cite{lee2021bbam}, point annotations~\cite{bearman2016s}, and image-level labels~\cite{lee2021anti}, to achieve pixel-level segmentation. This paper explores techniques based on image-level labels, which are particularly advantageous due to their ease of collection from internet sources and minimal annotation costs. 
    
    Current image-level WSSS techniques commonly include the following steps: (1) generating CAMs~\cite{zhou2016learning,wang2020self,xu2022multi} for specific categories leveraging a classification network to locate objects roughly; (2) refining them into pseudo-mask annotations~\cite{ahn2018learning}; (3) applying these pseudo-mask annotations and the original images to train a semantic segmentation network. Acquiring high-quality CAMs is vital for the subsequent processes~\cite{cheng2023out}. However, existing WSSS methods often suffer from inaccurate CAMs, such as the mistaken activation of non-target objects (object co-occurrence) and incomplete activation of target objects (under-activation), due to the limited semantic depth of image-level labels. These challenges hinder accurate localization and segmentation, especially in multi-target scenes. As shown in Figure~\ref{Figure 1}, this phenomenon occurs primarily due to the failure to fully capture complex interactions between deep structures and features while establishing a strong relationship between activation responses and image semantics. Prior knowledge of an object’s category can offer insights into its holistic features when interacting with the original image. Although this information may appear as simple words or numbers in low-dimensional space, it can be articulated in a more comprehensive and intricate manner in high-dimensional space.

    \begin{figure}[t]
    \centering
    \includegraphics[width=\columnwidth]{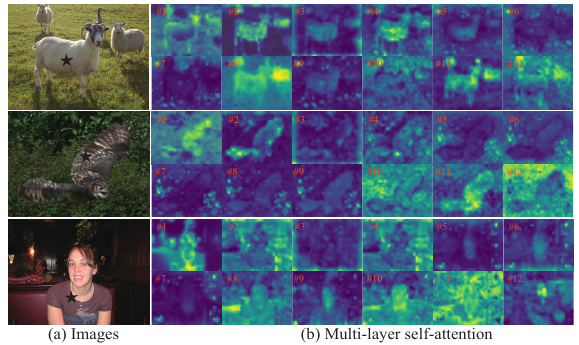}  %
    \caption{(a) The image and query targets (\(\star\)). (b) The self-attention maps in the Transformer block capture semantic-level relationships at various granularities. The high-activation regions learned by each layer not only provide critical information that subsequent layers may miss but also generate CAMs that often focus on different regions. This broadens the coverage of target feature areas, effectively mitigating the tendency of activation maps to focus excessively on local salient regions.}
    \label{Figure 2}
    \end{figure}
    
    Existing works predominantly emphasize extracting feature information for classification from input images, neglecting the critical role of high-dimensional semantic space in generating accurate CAMs~\cite{wang2020self}. This oversight typically results in CAMs that either cover only a portion of objects’ distinctive features or erroneously include non-target objects. As application scenarios for WSSS tasks grow more complex, limitations in the model’s ability to fully recognize contours and accurately locate target objects have become increasingly apparent. Our observations reveal that, in WSSS tasks, the cascaded encoders in ViT~\cite{dosovitskiy2021image}-based classification or representation learning models facilitate hierarchical and long-range information modeling for class-specific activations, as illustrated in Figure~\ref{Figure 2}. Inspired by this, we incorporate feedback from attention mechanisms across cascaded encoders into the embedding process to strengthen the influence of embeddings with reliable semantic information during information interaction. This enables class-specific feature learning at each layer. Additionally, by mitigating the occurrence of false positives through reducing the influence of unreliable embeddings, the final embedding retains the maximal useful information about the target object, thereby amplifying the model’s discriminative capability in generating CAMs.

    Furthermore, to improve the accuracy of feature representations in the multi-dimensional space during optimization, we propose a hybrid-feature alignment module that integrates RGB information of the original image with the embedding's cosine similarity features. The incorporation of this module aims to address the limitations of embedding representation capability in low-dimensional space. By employing this strategy, we further refine the model’s comprehension and representation of image semantics, leading to higher-quality object localization maps. Our main contributions are summarized as follows:
    
    \begin{itemize}
    
    \item We propose a novel mechanism, namely Dual Optimization of Embedding Information (DOEI), which is plug-and-play and can be applied to each layer encoder of the ViT. This mechanism effectively boosts useful information in the embedding while suppressing irrelevant information, achieving a rich representation of image features and intra-class diversity, thereby improving the accuracy of CAMs and reducing activation noise.
    
    \item We introduce a novel feature alignment module as a complementary optimization to DOEI. This module integrates the RGB values of the original image, the spatial features of embeddings, and self-attention scores, making the candidate tokens more meaningful, thereby facilitating the accurate representation of semantic structures and the effective transmission of embedding information.
    
    \item We incorporate the proposed mechanism into ViT-based models, with experiments demonstrating that this mechanism advances baseline model performance across different datasets without adding additional learnable parameters. It effectively prevents incorrect target localization and significantly improves the integrity of target recognition.
    \end{itemize}
    
\section{Related Work}

    \begin{figure*}[t] 
    \centering
    \includegraphics[width=\textwidth]{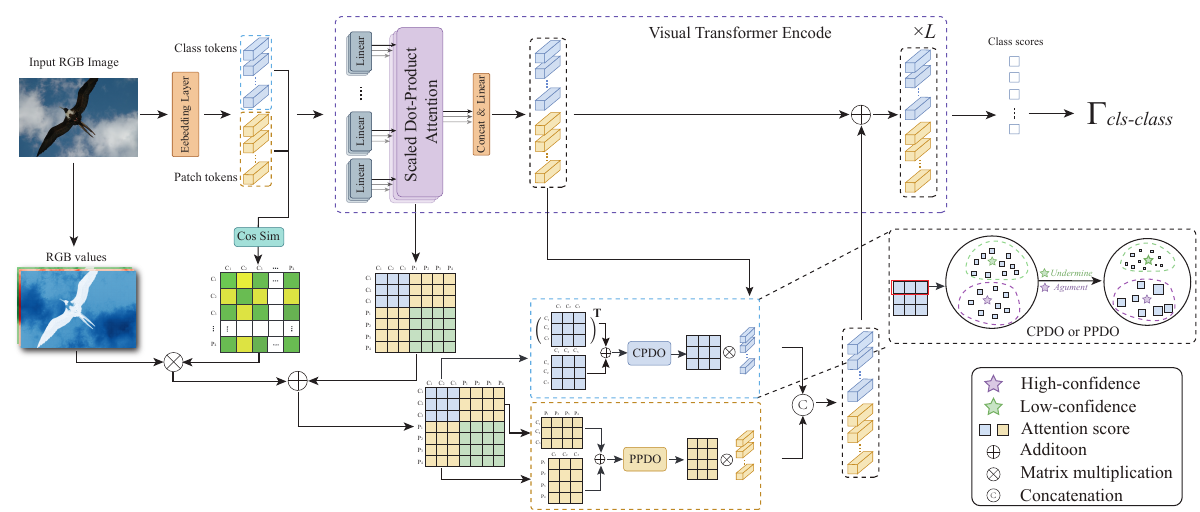}
    \caption{An overview of the proposed method. The RGB image undergoes transformation into class tokens and patch tokens via the embedding layer. The dot-product attention mechanism is then employed to compute the attention score matrix and generate output tokens. To further refine the attention score matrix, the cosine similarity between the RGB values of the original image and the tokens is used to adjust the distribution of attention weights. Furthermore, the CPDO and PPDO methods are customized to amplify high-confidence information and suppress the influence of low-confidence information. Finally, the optimized tokens are incorporated into the original tokens as residuals, producing refined output tokens for subsequent computations in the encoder.
    }
    \label{Figure 3} 
    \end{figure*}

\subsection{Weakly Supervised Semantic Segmentation}

    In WSSS, full pixel-level segmentation relies on limited supervisory signals. The development of WSSS has significantly alleviated the dependency on large amounts of pixel-level labels in traditional semantic segmentation models. Prevailing approaches mainly use the CAM technique introduced by Zhou et al.~[\citeyear{zhou2016learning}] as an initial step in identifying target object locations. The CAM combines the global average pooling layer with the classification layer to efficiently consolidate feature information for each pixel, generating an activation map that is aligned with the semantic representation of a specific category and serving as a key component of WSSS. However, CAMs typically display only the salient regions of the target object, hindering the capture of complete location information. Therefore, directly adopting CAM as full pixel-level segmentation labels is constrained by these shortcomings. Most of the research focuses on generating more precise and comprehensive CAMs and refining pseudo-labels to augment the fidelity and precision of segmentation results. By optimizing the CAM generation process and incorporating richer semantic information, researchers aim to mitigate issues of insufficient or overly concentrated activation areas, paving the way for more accurate and versatile WSSS methodologies.

\subsection{Generating High-quality CAMs}

    Convolutional Neural Networks (CNNs) \cite{qi2025medconv} and Vision Transformers (ViTs) \cite{wu2024xlip,ji2024sine,zhang2024jointvit} are two popular approaches for generating CAMs. CNN-based methods often face challenges with localized activation due to the limited receptive field and reliance on local features. To address this, strategies such as random occlusion~\cite{kumar2017hide} (e.g., "hide-and-seek") and adversarial training~\cite{kweon2023weakly} have been employed to expand activation areas. Techniques like dilated convolutions~\cite{huang2018weakly} and multi-layer feature integration~\cite{li2022weakly} further elevate segmentation accuracy by capturing more contextual information. Additionally, specialized loss functions, such as contrastive loss~\cite{zhu2024weakclip} and SEC loss~\cite{wu2022adaptive}, and supplementary data, including saliency maps and videos~\cite{wang2018weakly}, have been used to improve object localization.An alternative method for generating CAMs is to utilize the self-attention weight matrix in ViT~\cite{dosovitskiy2021image}. For example, Gao et al.~[\citeyear{gao2021ts}] combine semantic-aware annotations with semantically unrelated attention maps, providing a feasible approach for object localization by utilizing the semantic and localization information extracted by ViT. Ru et al.~[\citeyear{ru2022learning}] refine the initial pseudo-labels for segmentation by learning robust semantic affinities with the aid of a multi-head attention mechanism. Xu et al.~[\citeyear{xu2023learning}] transform simple class labels into high-dimensional semantic information through Contrastive Language-Image Pretraining (CLIP) to guide the ViT, forming a multimodal representation of text and images, thus generating more precise object localization maps.
    
\subsection{Refinement of CAMs}

     At present, existing refinement methods primarily focus on the second stage of multi-stage models. For instance, Ahn et al.~[\citeyear{ahn2018learning}] train AffinityNet by leveraging reliable foreground and background activation maps to predict affinities between neighboring pixels, which are then used as metrics for the random walk algorithm, thereby expanding the CAM. The IRN method~\cite{ahn2019weakly} further refines CAM by estimating object boundary information through the semantic affinities between the original pixels in the image. Wang et al.~[\citeyear{wang2020weakly}] propose a method that refines CAM by leveraging high-confidence pixels from segmentation results as inputs to a pairwise affinity network. Xu et al.~[\citeyear{xu2021leveraging}] highlight that affinities in saliency maps and segmented representations more effectively reflect the diversity in CAM representations. For WSSS tasks, such refinement is crucial for improving the final segmentation performance and helps generate more accurate and reliable pixel-level pseudo-labels.

\section{Methodology}

    \subsection{Preliminaries}

    For ViT-based multi-classification networks, an input image is first split into \( N \times N \) patches through a convolutional layer or a fully connected layer, which are then converted into \( N ^2 \) patch tokens \(\left\{ t_n \in \mathbb{R}^{1 \times D}, \, n = 1, 2, \dots, N^2 \right\}\), where \( D \) represents the embedding dimension of each token.  Inspired by MCTformer~\cite{xu2022multi}, among these patch tokens, \( C \) class tokens \( t_* \in \mathbb{R}^{C \times D} \) is concatenated, where \( C \) represents the number of classes, and \( t_* \) is a learnable parameter initialized randomly. After adding positional encoding, these tokens \( \Gamma \in \mathbb{R}^{ (N^2 + C) \times D} \) are fed into \( L \) cascaded encoders of the standard Transformer modules, where \( \Gamma \) represents the combination of \( t_n \) and \( t_* \). Specifically, tokens \( \Gamma \) are projected by employing three learnable parameter matrices \( W^Q \), \( W^K \), \( W^V \), which map them into query \( Q\in \mathbb{R}^{ (N^2 + C) \times d_q} \), key \( K \in \mathbb{R}^{ (N^2 + C) \times d_k} \), and value \( V \in \mathbb{R}^{ (N^2 + C) \times d_v} \). Here, \( d_q \) = \( d_k \), denotes the feature dimension for the query and key, while \( d_v \) indicates the feature dimension of the value. Subsequently, the scaled dot-product attention mechanism is applied to compute the attention matrix \( A \) between the query and key:

    \begin{equation}
    S_i = \Gamma_i W^{s}_i, \quad S \in \{Q, K, V\}, \text{and} \quad A_i = \frac{Q_i K_i^\top}{\sqrt{d_k}}
    \end{equation}

    \noindent where \( i \in \{1, 2, \dots, L\} \) denotes the \( i \)-th transformer block. The softmax function is applied to matrix \( A_i \), which is then multiplied by the key \( V_i \), after which a residual connection is added, followed by Layer Normalization and input into the MLP to produce output \( X_i \).

    \begin{equation}
    X_i = \text{MLP}(\text{LN}(\text{softmax}(A_i) V_i+ X_{i-1})) 
    \end{equation}
    
    \noindent where \( X_i \) denotes the output of the \( i \)-th transformer block. When \( i \) = \( L \), class tokens \( \Gamma_{cls} \in \mathbb{R}^{ C \times D } \) are extracted from \( X_i \) and averaged across channels to yield category scores \( y_{cls} \in \mathbb{R}^{ C } \). Subsequently, these category scores are matched with image-level ground truth labels via multi-label soft margin loss (\(MLSM \)), and the loss is calculated to enable supervised learning: 
    
    \begin{align}
    \mathcal{L}_{cls} &= \text{MLSM}(\mathbf{y}_{cls}, \mathbf{y}) \nonumber\\
    &= - \frac{1}{C} \sum_{i=1}^{C} y^i \log \sigma(y^i_{cls}) \\
    &\quad + (1 - y^i) \log (1 - \sigma(y^i_{cls})) \nonumber
    \end{align}
    
    \subsection{Overall Pipeline}
    
    Inspired by the modeling of long-range dependencies at different levels by the encoder layers in the standard ViT, we explore whether multi-scale inter-object coupling interactions can be utilized to optimize input embeddings during forward propagation, thereby extending the semantic diversity of anchor class activation features in the high-level semantic space. The overall architecture of DOEI is shown in Figure~\ref{Figure 3}. We employ TS-CAM~\cite{gao2021ts}, TransCAM~\cite{li2023transcam}, and MCTformer~\cite{xu2022multi} as baseline models to assess the effectiveness of the proposed method. In each baseline network, we adopt a standard multi-stage WSSS process: 1) Generating Class Activation Maps (CAMs); 2) Refining the CAMs; 3) Training the segmentation network.

    \textbf{Generating CAMs.} We generate CAMs by training a multi-class classifier on the baseline network. Specifically, the feature map output \( F \in \mathbb{R}^{ h \times w \times d} \) by the baseline network is weighted and summed with the weight matrix \( W \) of the classifier’s final layer to extract the corresponding class-specific feature map \( M^c \), followed by normalization. Finally, the CAM is generated by calculating the contrast threshold \( \beta \) \( (0 < \beta < 1) \) relative to the background.

    \begin{equation}
    M_{tmp}^c = \text{ReLu}\left( \sum_{i=1}^{d} W^{i,c} F^i \right) 
    \end{equation}
    
    \begin{equation}
    M^{i,j} = \begin{cases}
    \arg\max(M^{i,j,:}), & \text{if } \max(M_{tmp}^{i,j,:}) \geq \beta, \\
    0, & \text{if } \max(M_{tmp}^{i,j,:}) \leq \beta
    \end{cases}
    \end{equation}
    
    \noindent where ReLu is a nonlinear activation function used to filter out negative values in \( M_{tmp}^c \). Min-Max normalization is applied to scale \( M_{tmp}^c \) to [0, 1]. The pixel (\( i \), \( j \)) represents a specific location in the feature map.
    
    \textbf{Refining CAMs.} We employed the same method as the baseline network, utilizing the AffinityNet~\cite{ahn2018learning} introduced by Ahn et al. to refine the initial seeds. AffinityNet learns reliable affinities between pixels from the initial localization map as a supervisory signal and predicts an affinity matrix to enable stable expansion of seed regions, thereby generating pseudo-mask labels.
    
    \textbf{Training Segmentation Network.} Building on previous mainstream research~\cite{zhang2021complementary} and the baseline network configuration, we selected DeepLabv1~\cite{chen2014semantic} with a ResNet38~\cite{wu2019wider} backbone as the segmentation network. The segmentation network uses pseudo-mask labels as supervisory signals to perform regression prediction for each pixel. The loss function is given as: 
    
    \begin{equation}
    \mathcal{L}_{\text{seg}} = \frac{1}{N} \sum_{i=1}^{N} \sum_{j=1}^{C} -\hat{y}_i^j \log \left( \frac{\exp(z_i^j)}{\sum_{k=1}^{C} \exp(z_i^k)} \right) 
    \end{equation}
    
    \noindent where \( N \) represents the total number of pixels in the image; \( i \) denotes the index of each pixel, ranging from 1 to \( N \);and \( j \) represents the class index, ranging from 1 to \( C \). \(\hat{y}_i^j\) denotes the ground truth label for pixel \( i \), otherwise 0 (i.e., one-hot encoding).\( z_i^j \) denotes the logit value of the network output for pixel \( i \) belonging to class \( j \).
    
    \subsection{Dual Optimization of Embedding Information}

    Given the importance of high-quality CAMs for WSSS, this paper focuses on CAMs generation. Previous research has mainly focused on extracting key classification features from the image, frequently overlooking the importance of guiding information generated through the interaction between semantic content and image features. Therefore, we propose a dual optimization mechanism for embedding information (DOEI) in ViT to fully utilize the interaction between semantic content and image features. The proposed DOEI consists of two independent modules: Class-wise Progressive Decoupling Optimization (PPDO) and Patch-wise Progressive Decoupling Optimization (CPDO), which are used to optimize patch-to-class and class-to-class interactions, respectively.

    \textbf{Patch-wise Progressive Decoupling Optimization.} Traditional methods typically generate a weight matrix \( A_i \) through the dot-product self-attention mechanism of class tokens and patch tokens in the \( i \)-th layer, and multiply it by the key generated from the embedding of the input in the \( (i-1) \)-th layer to obtain the input embedding for the \( (i+1) \)-th layer. The \( S_{p2c}^i \), where \( S_{p2c}^{i,x} = A_i[1 : C, C + 1 : C + N^2] \in \mathbb{R}^{ N^2 \times C} \) represents the attention score of patch-to-class, and \( S_{p2c}^{i,y} = A_i[C + 1 : C + N^2, 1 : C + 1] \in \mathbb{R}^{ N^2 \times C} \) represents the attention score of class-to-patch, with \(x\) and \(y\) used here to denote the two different forms of \( S_{p2c}^{i}\). These scores reflects the degree of relevance of the image information in the patch to that class. In other words, the \( S_{p2c} \) indicates the most likely class affiliation for that patch. Unreliable \( S_{p2c} \) (e.g., when a patch lacks relevant positional information for the class) can result in the creation of numerous false-positive pixels in the CAM during subsequent embedding transmission, causing incorrect class activation. Furthermore, assigning greater weight to high-confidence \( S_{p2c} \) can amplify their influence in subsequent embeddings, thereby guiding the model to focus more on specific categories. We partitioned all patch-to-class attention scores \( S_{p2c} \) into Patch Candidate Confidence Score (\(PC_{cs}\)) and Patch Candidate Non-confidence Scores (\(PC_{ns}\)):

    \begin{equation}
    \mathcal{P} = S_{p2c} = S_{p2c}^{x} + (S_{p2c}^{y})^T
    \end{equation}

    \begin{equation}
    PC_{cs} = \bigcup_{i=1}^{m} \left\{ a_{i,j} \mid j \in \operatorname{TopIndices}\left( \mathcal{P}_{i,j}, t \right) \right\}
    \end{equation}

    \begin{equation}
    PC_{ns} = \bigcup_{i=1}^{m} \left\{ b_{i,j} \mid j \in \operatorname{BottomIndices}(\mathcal{P}_{i,j}, 1-t) \right\}
    \end{equation}

    \noindent denote \(m\) and \(n\) as the dimensions of \(\mathcal{P}\), where \(m\) indicates the number of classes and \(n\) indicates the number of patches. \(t\) is calculated by the following formula:

    \begin{equation}
    t = n*(L - i)*ST_{p2c}       
    \end{equation}

    \noindent where \( i \in \{1, 2, \dots, L\} \) represents the \( l \)-th transformer block and \(L\) represents the total number of transformer layers. The selective threhold (\(ST_{p2c}\)) represents a hyperparameter used to control the division threshold. We consider that each layer of the self-attention mechanism learns image information at different scales. Therefore, we employed a progressive strategy that selects different numbers of confidence and non-confidence scores based on the features of each layer.

    We apply enhancement and suppression operations to \(PC_{cs}\) and \(PC_{ns}\), respectively, calculate the optimized embedding vectors, and combine them with the original embedding vectors with residual connections:

    \begin{align}
    X^p_i &= X_i[m:,:] \nonumber + X_i[n:m,:] \times A_i[PC_{cs}] * AF_{p2c} \\
        &+ X_i[m:,:] \times A_i[PC_{ns}] * SF_{p2c}   
    \end{align}

    \noindent where the Augment Factor (\(AF_{p2c}\)) and Suppression Factor (\(SF_{p2c}\)) are controllable hyper-parameters.

    \textbf{Class-wise Progressive Decoupling Optimization.} Class-to-class attention scores capture the similarity among categories. In multi-object, complex scenes, where the model relies solely on image-level annotations, it often struggles to accurately capture target location information, making it difficult to align these locations with semantic data. To address this challenge, CPDO is introduced to establish complementary similarities between categories. Specifically, the location information of one category can be supplemented by the semantic information of other similar categories, thereby enhancing the model's ability to perceive target locations and reinforcing the correspondence between semantic content and spatial positions. Similarity, we partitioned all class-to-class attention scores \( S_{c2c} \) into Class Candidate Confidence Score \(CC_{cs}\) and Class Candidate Non-confidence Score \(CC_{us}\):Similarly, we divide all class-to-class attention scores \( S_{c2c} \) into candidate confidence scores \(CC_{cs}\) and non-confidence scores \(CC_{us}\),then apply enhancement and suppression operations to obtain the optimized embedding vectors \(X^c_i\):

    \begin{equation}
    \mathcal{C} = S_{c2c} = S_{c2c}^{x} + (S_{c2c}^{y})^T
    \end{equation}

    \begin{equation}
    C_{cs} = \bigcup_{i=1}^{n} \left\{ a_{i,j} \mid j \in \operatorname{TopIndices}\left( \mathcal{C}_{i,j}, t \right) \right\}
    \end{equation}

    \begin{equation}
    C_{ns} = \bigcup_{i=1}^{n} \left\{ b_{i,j} \mid j \in \operatorname{BottomIndices}(\mathcal{C}_{i,j}, 1-t) \right\}
    \end{equation}

    \begin{equation}
    t = n*(L - i)*ST_{c2c}       
    \end{equation}

    \begin{align}
    X^c_i &= X_i[n:m,:] \nonumber + X_i[n:m,:] \times A_i[CC_{cs}] * AF_{c2c} \\
        &+ X_i[n:m,:] \times A_i[CC_{ns}] * SF_{c2c}   
    \end{align}

    Finally, \(X^p_i\) and \(X^c_i\) are concatenated to form a new embedding vector \(X_{i+1}\), which is then fed as input to the next layer of the self-attention mechanism.
    
    \subsection{Hybrid Feature Alignment Module}

    The self-attention scores, which fully represent image feature information, provide a reliable foundation for embedding optimization in DOEI. Relying solely on the attention weight matrix generated by the ViT encoder as a source of potential information for embedding optimization is insufficient to fully capture the high-dimensional features of objects anchored in the image. Therefore, to comprehensively capture the latent semantic and structural features within the image, the process of selecting feature information needs to be re-evaluated to encompass a broader range of rich content. This paper introduces a novel feature information construction method that effectively extracts key information from images by integrating self-attention weights, raw RGB values, and features from the ViT patch embedding layer. The specific construction method is as follows:

    \begin{equation}
    A = (1 - \alpha) \cdot W_{\text{attn}} + \alpha \cdot d_{\text{rgb}}(x) \cdot d_{\text{emb}}(X) 
    \end{equation}

    \noindent here, \( W_{attn}\) represents the self-attention weight matrix generated by the ViT model. Denote \( d_{rgb}(x) \) and \( d_{emb}(X) \) as the normalized pure RGB features of the original image and the cosine similarity based on Transformer embedding features, respectively.
    
\section{Experiments}

    \subsection{Datasets and Evaluation Protocol} The datasets employed in the experiments -- PASCAL VOC 2012~\cite{everingham2010pascal} and MS COCO 2014~\cite{lin2014microsoft} -- along with a detailed description of the evaluation metrics, are provided in \textbf{Supplementary Section 1}.

    \subsection{Implementation Details}

    We integrate the proposed method in three baseline networks (\(i.e\)., MCTformer~\cite{xu2022multi}, TS-CAM~\cite{gao2021ts}, TransCAM~\cite{li2023transcam}). All classification networks for generating CAMs use the parameter settings of the original baseline. Due to differences in network architecture, the Patch-wise Progressive Decoupling Optimization (PPDO) module is only applied in TransCAM and TS-CAM. We draw on previous work~\cite{xu2022multi} and apply PSA~\cite{ahn2018learning} to the generated initial CAMs to obtain pseudo-mask labels (Mask). The final semantic segmentation network uses DeepLabV1~\cite{chen2014semantic} based on ResNet38~\cite{wu2019wider}, with parameter settings consistent with the compared baseline methods. In testing, we evaluate the model performance with inputs at multiple scales (0.5, 0.75, 1.0, 1.25, 1.5) and apply DenseCRFs~\cite{chen2014semantic} post-processing to the output results.
    
    \subsection{Main Results}

    \textbf{Pascal VOC.} In Table ~\ref{Table 1}, we demonstrate the improvement in seed object localization maps after incorporating the DOEI mechanism into baseline methods, and compare these results with those of classical WSSS methods. As shown in Table ~\ref{Table 1}, the baseline models with DOEI achieved improvements of 6.8\%, 1.4\%, and 3.8\% on seed, and enhancements of 4.7\%, 1.1\%, and 3.7\% on pseudo-mask labels (mask) after PSA processing. Additionally, Table  ~\ref{Table 2} presents the performance of segmentation models trained with pseudo-labels generated by the baseline models incorporating DOEI on the Pascal VOC validation and test sets. The results show that the proposed mechanism significantly improved the baseline models, with performance increases of 4.1\%, 0.9\%, and 0.9\%, respectively.

    \begin{table}[H]
    \fontsize{8pt}{10pt}\selectfont
    \centering
    \begin{tabular}{lcc}
        \toprule
        Method  & Seed & Mask \\
        \midrule
        AdvCAM~[\citeyear{lee2021anti}] & 55.6 & 69.9 \\
        CDA~[\citeyear{su2021context}] &  55.4  & 63.4\\
        CPN~[\citeyear{zhang2021complementary}] & 57.4 & 67.8\\
        ReCAM~[\citeyear{chen2022class}] & 54.8 & 69.7 \\
        CLIMS~[\citeyear{xie2022clims}] & 56.6 & 70.5 \\
        FPR~[\citeyear{chen2023fpr}] & 60.3 & - \\
        SFC~[\citeyear{zhao2024sfc}] & 64.7 & 73.7 \\
        \midrule
        TS-CAM~[\citeyear{gao2021ts}] & 29.9 & $41.4^{\star}$ \\
        \rowcolor{lightblue} DOEI+TS-CAM & \textbf{36.7} \textcolor{red}{↑ \textbf{6.8}}& \textbf{45.5} \textcolor{red}{↑ \textbf{4.1}} \\
        \cdashline{1-3}
        TransCAM~[\citeyear{li2023transcam}]  & 64.9 & 70.2\\
        \rowcolor{lightblue} DOEI+TransCAM & \textbf{66.3} \textcolor{red}{↑ \textbf{1.4}}& \textbf{71.3} \textcolor{red}{↑ \textbf{1.1}}\\
        \cdashline{1-3}
        MCTformer~[\citeyear{xu2022multi}] & 61.7 & 69.1 \\
        \rowcolor{lightblue} DOEI+MCTformer  & \textbf{65.5} \textcolor{red}{↑ \textbf{3.8}}& \textbf{72.2} \textcolor{red}{↑ \textbf{3.7}}\\
        \bottomrule
    \end{tabular}
    \caption{Evaluation of the initial seed (Seed) and its corresponding pseudo segmentation ground-truth mask (Mask) is conducted through mIoU (\%) on the PASCAL VOC \(train\) set. \(\star\) denotes our reproduced result.}
    \label{Table 1}
    \end{table}
    
    \begin{table}[H]
    \fontsize{8pt}{10pt}\selectfont
    \centering
    \begin{tabular}{lccc}
        \toprule
        Method  & Backbone & Val & Test \\
        \midrule
        SEAM~[\citeyear{wang2020self}]  & ResNet101 & 64.5 & 65.7\\
        AuxSegNet~[\citeyear{xu2021leveraging}] & ResNet38 & 69.0 & 68.6\\
        EPS~[\citeyear{lee2021railroad}] & ResNet101 & 71.0 & 71.8\\
        ECS-Net~[\citeyear{sun2021ecs}] & ResNet38 & 64.3 & 65.3\\
        AdvCAM~[\citeyear{lee2021anti}] & ResNet101 & 68.1 & 68.0\\
        CDA~[\citeyear{sun2021ecs}] & ResNet38 & 66.1 & 66.8\\
        Spatial-BCE~[\citeyear{wu2022adaptive}] & ResNet38 & 70.0 & 71.3\\
        ReCAM~[\citeyear{chen2022class}] & ResNet101 & 69.5 & 69.6\\
        SIPE~[\citeyear{chen2022self}] & ResNet101 & 68.8 & 69.7 \\
        CLIMS~[\citeyear{xie2022clims}] & ResNet101 & 70.4 & 70.0\\
        LPCAM~[\citeyear{chen2023extracting}] & ResNet101 & 71.8 & 72.1\\
        CLIP-ES~[\citeyear{lin2023clip}] & ResNet101 & 73.8 & 73.9\\
        \midrule
        TS-CAM~[\citeyear{gao2021ts}] & ResNet38 & $43.2^{\star}$ & $42.9^{\star}$ \\
        \rowcolor{lightblue} DOEI+TS-CAM & ResNet38 & \textbf{46.8} \textcolor{red}{↑ \textbf{3.6}}& \textbf{47.0} \textcolor{red}{↑ \textbf{4.1}} \\
        \cdashline{1-3}
        TransCAM~[\citeyear{li2023transcam}]  & ResNet38 & 69.3 & 69.6\\
        \rowcolor{lightblue} DOEI+TransCAM & ResNet38 & \textbf{70.8} \textcolor{red}{↑ \textbf{1.5}}& \textbf{70.5} \textcolor{red}{↑ \textbf{0.9}}\\
        \cdashline{1-3}
        MCTformer~[\citeyear{xu2022multi}] & ResNet38 & 70.2\textcolor{black}{†} & 70.1\textcolor{black}{†} \\
        \rowcolor{lightblue} DOEI+MCTformer  & ResNet38 & \textbf{71.4} \textcolor{red}{↑ \textbf{1.2}}& \textbf{71.0} \textcolor{red}{↑ \textbf{0.9}}\\
        \bottomrule
    \end{tabular}
    \caption{Performance comparison of WSSS methods in terms of mIoU (\%) on the PASCAL VOC \(val\) and \(test\) split through different segmentation backbones. †: Our reimplemented results by means of official code. Note that TS-CAM doesn’t provide official evaluation results on PASCAL VOC dataset, the results of TS-CAM (\(\star\)) are derived by us.}
    \label{Table 2}
    \end{table}
    
    \textbf{MS COCO.} Table~\ref{Table 3} reports the results of our DOEI compared to previous methods on MS COCO \(val\) split. MS COCO is a complex dataset containing 80 classification categories, covering a wide range of objects from everyday life. Our method achieved performance improvements of 1.2\% and 1.6\% over the baseline model, respectively.

    \subsection{Ablation Study and Parameter Analysis}

    We perform ablation and parameter experiments on the PASCAL VOC \(train\) set to assess the effectiveness of the proposed method. All experiments are conducted via MCTformer~\cite{xu2022multi}, with DeiT-S serving as the backbone network.

    \begin{table}[H]
    \fontsize{8pt}{10pt}\selectfont
    \centering
    \begin{tabular}{lcc}
        \toprule
        Method  & Backbone & Val \\
        \midrule
        CONTA~[\citeyear{zhang2020causal}] & ResNet38 & 32.8\\
        AuxSegNet~[\citeyear{xu2021leveraging}] & ResNet38 & 33.9 \\
        CDA~[\citeyear{sun2021ecs}] & ResNet38  & 33.2\\
        EPS~[\citeyear{lee2021railroad}] & ResNet101 & 35.7\\
        ReCAM~[\citeyear{chen2022class}] & ResNet38  & 42.9\\
        SIPE~[\citeyear{chen2022self}] & ResNet101 & 43.6\\
        \midrule
        TransCAM~[\citeyear{li2023transcam}]  & ResNet38 & $43.5^{\star}$ \\
        \rowcolor{lightblue} DOEI+TransCAM & ResNet38 & \textbf{44.7} \textcolor{red}{↑ \textbf{1.2}}\\
        \cdashline{1-3}
        MCTformer~[\citeyear{xu2022multi}] & ResNet38 & 42.0 \\
        \rowcolor{lightblue} DOEI+MCTformer  & ResNet38 & \textbf{43.6} \textcolor{red}{\textbf{↑ 1.6}}\\
        \bottomrule
    \end{tabular}
    \caption{Performance comparison of WSSS methods in terms of mIoU (\%) on the MS COCO \(val\) set. Note that TransCAM doesn’t provide official evaluation results on MS COCO dataset, the results of TransCAM (\(\star\)) are implemented by us.}
    \label{Table 3}
    \end{table}

    \textbf{The Effectiveness of the Components.} Table~\ref{Table 4} presents the performance improvements contributed by each component of the Dual Optimization of Embedding Information (DOEI) mechanism and the Hybrid-Feature Alignment (HFA) module. Introducing either Patch-wise Progressive Decoupling Optimization (PPDO) or Class-wise Progressive Decoupling Optimization (CPDO) individually increases the model's seed accuracy from 61.7\% to 64.2\% and 64.3\%, respectively. Combining PPDO and CPDO to form the DOEI mechanism yields a significant performance improvement, achieving an mIoU of 65.1\%. Further integrating DOEI with HFA results in an even greater performance boost, with an mIoU of 65.5\%. These results highlight the DOEI method's ability to enhance the global representation of target objects while effectively suppressing interference from non-target objects. Additionally, the findings confirm the effectiveness of the HFA module in addressing the limitations of relying solely on attention scores as features, thereby enhancing overall model performance.

    \begin{table}[H]
    \fontsize{8pt}{10pt}\selectfont
    \centering
    \begin{tabular}{ccccc}
        \toprule
        Baseline & CPDO & PPDO & HFA & mIoU \\
        \midrule
        \checkmark &  &  &  & 61.7 \\
        \checkmark & \checkmark &  &  & 63.3 \\
        \checkmark &  & \checkmark &  & 63.2 \\
        \checkmark & \checkmark & \checkmark &  & 64.3 \\
        \checkmark & \checkmark &  & \checkmark & 64.6 \\
        \checkmark &  & \checkmark & \checkmark & 64.9 \\
        \checkmark & \checkmark & \checkmark & \checkmark & \textbf{65.5} \\
        \bottomrule
    \end{tabular}
    \caption{Performance improvements from different optimization mechanisms in the mIoU (\%) evaluation metric on the PASCAL VOC \(train\) set.}
    \label{Table 4}
    \end{table}
    
\begin{figure}[H]
    \centering
    \includegraphics[width=\columnwidth]{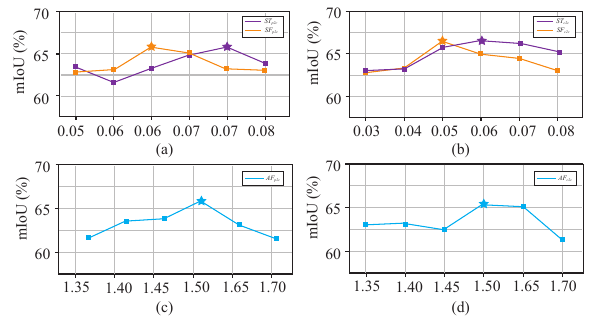}  
    \caption{The impact of different hyper-parameter values in the CPDO and PPDO modules on mIoU.}
    \label{Figure 4}
    \end{figure}

    \textbf{Hyper-parameters in PPDO and CPDO.} Figure~\ref{Figure 4} shows the results of analyzing each of the following parameters individually, while keeping the other hyper-parameters fixed: selection threshold, amplification factor, and suppress-factor for both patch-to-class and class-to-class (\( AF_{p2c} \), \( SF_{p2c} \) and \( ST_{c2c} \), \( AF_{c2c} \), \( ST_{p2c} \)). These hyper-parameters are used to generate the optimal CAM configuration. The experimental results show that when these parameter values fall within a certain range, the mIoU achieves better values. As shown in Figure ~\ref{Figure 4}(a) and (c), by amplifying the patch information most likely to belong to a specific class (as learned by the model), the model improves its attention to that class, expanding the activation range; conversely, by suppressing the information least likely to belong to that class, the model reduces its focus on irrelevant regions, effectively preventing false activations. Additionally, class-to-class attention scores reflect the similarity of different semantic features. By leveraging the commonalities of similar semantics within the image, the model can selectively activate complementary semantic information, even in the absence of explicit class guidance. When an appropriately selected value for \( ST_{c2c} \), \( AF_{c2c} \) and \( SF_{c2c} \) are used, the model's embedding information can be optimized toward the correct semantic direction, thereby improving its performance and accuracy, as shown in Figure~\ref{Figure 4}(b) and (d).

    \textbf{Additional Parameter Analysis.} To verify the effectiveness of the proposed method, we also conduct experiments to analyze the parameters of the Hybrid Feature Alignment weight \( \alpha \) and the selection of embedding optimization layers. When the \( \alpha \) value falls within a specific range, the mIoU value significantly improves. Furthermore, as the number of optimization layers increases, the mIoU value reaches its maximum when the layer count is at its maximum, achieving optimal performance. The detailed experimental results are provided in \textbf{Supplementary Section 2}.

    \subsection{Qualitative Analysis}

    We provide qualitative comparison results of the baseline network (i.e., TS-CAM, TransCAM and MCTformer) with DOEI on representative examples from the PASCAL VOC and MS COCO datasets in \textbf{Supplementary Figure 2}. These comparisons provide a clear visual demonstration of the substantial improvements achieved by our method, particularly in enhancing the target localization accuracy of the baseline models. By refining embedded information, DOEI surpasses baseline approaches, demonstrating its ability to deliver more precise localization across diverse, challenging datasets.

\section{Conclusion}

    In this paper, we investigate the optimization of input embeddings by coupling class tokens and patch tokens across multiple scales of self-attention, thus enriching the diversity of anchored class activation features. We propose a Dual Optimization of Embedding Information mechanism (DOEI) that integrates seamlessly with the ViT. DOEI leverages coupled attention within the self-attention mechanism to amplify the semantic information of specific categories and suppress noise during forward propagation, effectively reconstructing the semantic representation of embeddings. Additionally, we construct hybrid feature representations by combining the RGB values of the image, embedding features, and self-attention scores, thereby enhancing the reliability of candidate tokens within the DOEI mechanism. Experimental results show that the baseline models with DOEI successfully alleviate over-activation and under-activation issues on the Pascal VOC and MS COCO datasets, greatly improving the quality of class activation maps and boosting semantic segmentation performance.
    
\section*{Ethical Statement}

There are no ethical issues.

\clearpage

\newcommand{\supplementarytitle}{
    \twocolumn[
        \begin{center}
            \LARGE \textbf{DOEI: Dual Optimization of Embedding Information With Attention-Enhanced Class Activation Maps} \\
            \vspace{2em}
            \Large \textbf{- Supplementary Materials -}
            \vspace{4em}
        \end{center}
    ]
}

\supplementarytitle

\renewcommand{\thesection}{\arabic{section}}
\setcounter{section}{0}

\renewcommand{\thefigure}{\arabic{figure}}
\setcounter{figure}{0}

\renewcommand{\thetable}{\arabic{table}}
\setcounter{table}{0}

\section{Datasets and Evaluation protocol}

    We assess the effectiveness of the proposed method by evaluating its performance on the widely used datasets, PASCAL VOC 2012~\cite{everingham2010pascal} and MS COCO 2014~\cite{lin2014microsoft}. \textbf{PASCAL VOC} is divided into training (train), validation (val), and test sets, containing 1,464, 1,449, and 1,456 images, respectively, covering 20 target categories and one background category. To maintain consistency with the general protocol followed in previous studies~\cite{xu2022multi,cheng2023out}, we used the augmented training set containing 10,582 images to train the classification network. \textbf{MS COCO} includes 80 target categories and one background category, with approximately 82,000 training images and 40,000 validation images. Following~\cite{lee2021railroad}, we retained only images containing target classes and extracted the ground-truth labels from COCO stuff~\cite{caesar2018coco}.
    
    To ensure fair comparison of experimental results with previous studies~\cite{xu2022multi,cheng2023out}, we followed the same calculation  approach and used mean Intersection-over-Union (mIoU) as the evaluation metric, which is suitable for assessing the \(train\) and \(val\) sets of two datasets. For the PASCAL VOC \(test\) set, results are provided by the official online evaluation server.

    \begin{figure}[H]
    \centering
    \includegraphics[width=\columnwidth]{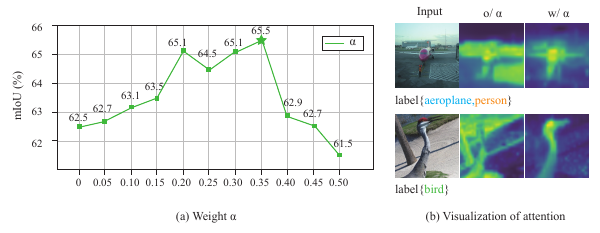}  
    \caption{Impact of varying numbers of embedding optimization layers on the mIoU, FP, and FN of the final CAM.}
    \label{Figure 5}
    \end{figure}
    
\section{Parameter Analysis}

    \textbf{Hybrid Feature Alignment weight \( \alpha \).} To highlight the impact of weight values \( \alpha \) on experimental results, all other hyper-parameters were fixed to the configuration that produced the best CAM quality. As shown in Figure~\ref{Figure 5}(a), the graph illustrates the effect of varying weights on mIoU values. Additionally, to visually demonstrate the improvement brought by the hybrid feature alignment weight \( \alpha \), we performed a comparative heatmap visualization of the fusion strategy.

    As shown in Figure~\ref{Figure 5}, panel (b) displays the heatmap visualizations of self-attention weights for a specific category, both without and with the introduction of hybrid feature alignment. Experimental results show that when the weight is set to 0.35, mIoU reaches its maximum value, improving by 0.7\% compared to the case without hybrid feature alignment. This demonstrates that the introduction of hybrid feature alignment module allows the model to effectively address the limitations of the ViT self-attention mechanism in capturing image features. The fused features offer a more comprehensive representation of information across various dimensional spaces in the image, thereby fully validating the effectiveness of the feature fusion strategy.
    
    \textbf{Selection of Embedding Optimization Layers.} To showcase the improvement in CAM generation from embedding optimization and identify the optimal number of embedding optimization layers, we optimized embeddings at different numbers of cascaded encoders and compared the resulting improvements in the target localization map quality. We use mIoU to measure the accuracy of CAM and introduce the false positive (FP) and false negative (FN) metrics to jointly assess the severity of over-activation and under-activation in the model. As shown in Figure~\ref{Figure 6}, we set different numbers \( K\) of embedding optimization layers across layers 1 to 12 (the transformer encoder in DeiT-S consists of 12 layers) and calculated the results separately for each configuration, where \( K\) refers to refers to the first \( K\) layers near the input end. The results indicate that embedding optimization at every layer most effectively suppresses over-activation or under-activation. In contrast, without embedding optimization, the FP and FN values are the highest, and the mIoU value is the lowest. This suggests that, with the help of the embedding optimization mechanism, the model can appropriately extend to the entire target area while suppressing extension to non-target regions.
    
    \begin{figure*}[t] 
    \centering
    \includegraphics[width=\textwidth, trim=0.1cm 0.1cm 0.1cm 0.1cm, clip]{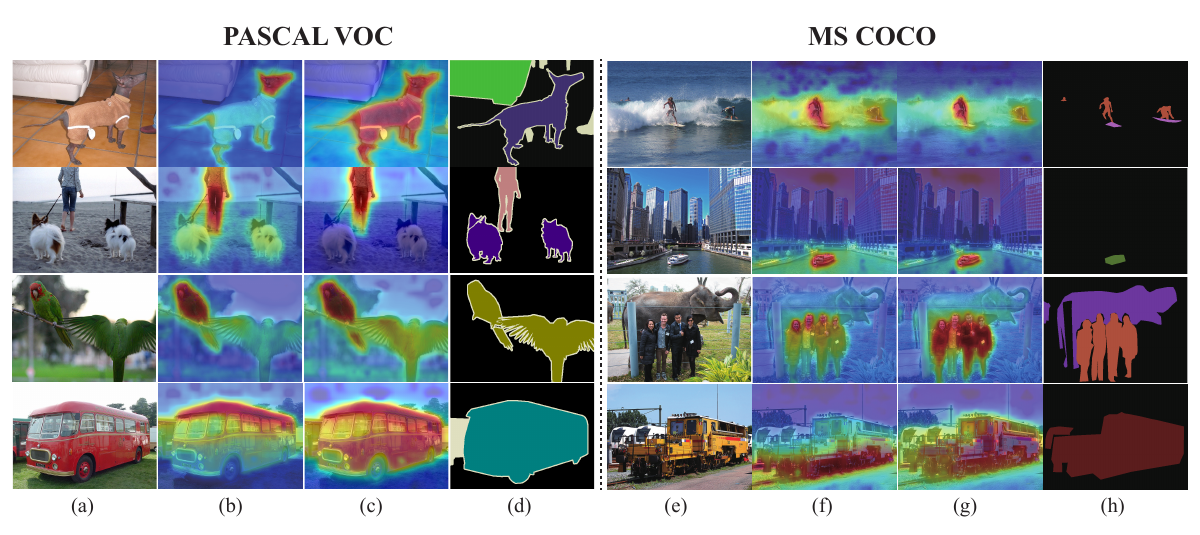}
    \caption{
    Visualization of the CAMs of input images generated by different methods. (a)(e) Input; (b)(f) Baseline; (c)(g) Dual Optimization of Embedding Information (DOEI); (d)(h) GT;
    }
    \label{Figure 7} 
    \end{figure*}

\section{Qualitative Analysis}

    Figure~\ref{Figure 7} illustrates a clear comparison of the results obtained by applying our proposed method versus those of the baseline model. On the PASCAL VOC dataset, the quality of class activation maps (CAMs) generated after optimizing with our information embedding technique is significantly superior to that of the original baseline model. Notably, even on large-scale and highly complex datasets such as MS COCO, our innovative method demonstrates outstanding and consistent performance. This is largely attributed to its ability to effectively extract, refine, and optimize the intrinsic knowledge learned by the model, enabling a more precise representation of the target features and progressively achieving enhanced overall performance.

    \begin{figure}[t]
    \centering
    \includegraphics[width=\columnwidth]{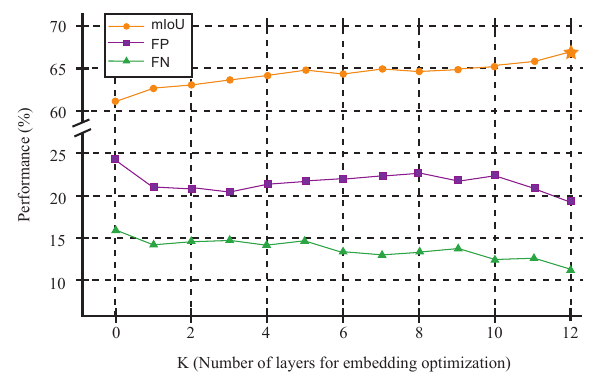}  
    \caption{Impact of varying numbers of embedding optimization layers on the mIoU, FP, and FN of the final CAM.}
    \label{Figure 6}
    \end{figure}

\end{document}